\documentclass[runningheads]{llncs}
\usepackage{graphicx}
\usepackage{comment}
\usepackage{amsmath,amssymb,amsfonts} % define this before the line numbering.
\usepackage{color}
\usepackage{times}
\usepackage{epsfig,graphicx}
\usepackage{algorithmic}
\usepackage{textcomp}
\usepackage{verbatim}
\usepackage{textcomp}
\usepackage{siunitx}

\usepackage{times,epsfig,graphicx}
\usepackage{algorithm,algorithmic}
\usepackage{amssymb,amsmath}
\usepackage{siunitx}

\makeatletter

\DeclareRobustCommand\onedot{\futurelet\@let@token\@onedot}
\def\@onedot{\ifx\@let@token.\else.\null\fi\xspace}

\def\be{\begin{equation}}
\def\ee{\end{equation}}
\def\bea{\begin{eqnarray}}
\def\eea{\end{eqnarray}}
\def\ben{\begin{eqnarray*}}
\def\een{\end{eqnarray*}}

\def\bi{\begin{itemize}}
\def\ei{\end{itemize}}

\newcommand{\bt}[1]{\begin{tabular}{#1}}
\newcommand{\et}{\end{tabular}}
\newcommand{\ba}[1]{\begin{array}{#1}}
\newcommand{\ea}{\end{array}}

\def\<{\langle}
\def\>{\rangle}

\newcommand{\hide}[1]{}

\newcommand{\bfx}{{\bf x}}

\newcommand{\bfq}{{\bf q}}

\newcommand{\bfd}{{\bf d}}

\newcommand{\bfp}{{\bf p}}
\newcommand{\bft}{{\bf t}}
\newcommand{\ra}{{\rightarrow}}

\usepackage[width=122mm,left=12mm,paperwidth=146mm,height=193mm,top=12mm,paperheight=217mm]{geometry}

\begin{document}
% \linenumbers
\pagestyle{headings}
\mainmatter

\title{Adversarial Transfer of Pose Estimation Regression}
%Adversarial Transfer of Pose Localization Models} 

\author{Boris Chidlovskii, Assem Sadek}
\institute{Naver Labs Europe, chemin Maupertuis 6, Meylan-38240, France
\email{firstname.lastname@naverlabs.com}\\
}
%\maketitle

%\begin{comment}
%\titlerunning{ECCV-20 submission ID \ECCVSubNumber} 
%\authorrunning{ECCV-20 submission ID \ECCVSubNumber} 
%\author{Anonymous ECCV submission}
%\institute{Paper ID \ECCVSubNumber}
%\end{comment}
%******************
\maketitle

%\begin{comment}
%\titlerunning{Adversarial Transfer...}
%\author{Albert Author\inst{1}\orcidID{0000-1111-2222-3333}}
%\end{comment}

%%%%%%%%%%%%%%%%%%%%%%%%%%%%%%%%%%%%%%%%%%%%%%%%%%%%%%

\begin{abstract}
We address the problem of camera pose estimation in visual localization. 
%The low generalization of 
Current regression-based methods for pose estimation are trained and evaluated scene-wise. They depend on the coordinate frame of the training dataset
and show a low generalization across scenes and datasets.
We identify the dataset shift an important barrier to generalization and consider transfer learning as an alternative way towards a better reuse of pose estimation models. 
%The task is defined as an adaptation of a pose regression models across scenes. 
%To benefit from %for in the image classification and semantic segmentation. 
We revise domain adaptation techniques for classification and extend them to camera pose estimation, which is a multi-regression task.
%with separate networks for learning the position and orientation of the pose.
%We consider the problem of representation transfer between two domains
%We extend domain adaptation techniques to the absolute pose regression (APR). 
We develop a deep adaptation network for learning scene-invariant image representations and use adversarial learning to generate such representations for model transfer. We enrich the network with self-supervised learning and use the adaptability theory to validate the existence of scene-invariant representation of images in two given scenes. We evaluate our network on two public datasets, Cambridge Landmarks and 7Scene, demonstrate its superiority over several baselines and compare to the state of the art methods.
%we propose an approach when a tiny part of the target instances are annotated to guide the model transfer. 
%We conclude the paper by a discussion about alternative methods including the unsupervised representation adaptation. 
%To guide the transfer, we consider several options of supplementary target information to guide the model transfer. 
%where the about target scene is given by small fraction is available poses.
%With such a supplement on the target poses, 
%We can learn scene-invariant representation of source and target images and transfer the source APR model to the target scene.l.
\end{abstract}

%-----------------------------------

\section{Introduction}
\label{sec:intro}

%----\cite{sattler_understanding_2019}
Visual localization is the task of accurate camera pose estimation in a known scene. It is a fundamental problem in robotics and computer vision, with multiple applications in autonomous vehicles, structure from motion (SfM), simultaneous localization and mapping (SLAM), Augmented Reality (AR) and Mixed Reality (MR)~\cite{shavit_introduction2019}. 

%Traditionally, the localization problem has been tackled using 3D geometry. Recently, end-to-end approaches based on convolutional neural networks have become popular. These methods learn to directly regress the camera pose from an input image.

%----- \cite{laskar_camera_2017}-------
%Due to importance of these problems various relocalization approaches have been proposed. 
%Research on visual localization was and still is dominated by structure-based methods. These 
Traditional structure-based methods find correspondences between local features extracted from an image by applying image descriptors like SIFT, SURF or ORB~\cite{leng2018local,rublee11orb} and 3D geometry of the scene obtained from SfM; obtained 2D-3D matches allow to recover the 6-DoF camera pose. 
%Unfortunately, due to low-level process of finding the matches, 
%structure-based methods do not work well 
%robustly and accurately in all scenarios, such as
%in scenes with large illumination changes, occlusions and repetitive structures~\cite{laskar_camera_2017,walch17image}.
% ---zhou2019learn -
%The 3D scene geometry used in the matching stage can
%be represented either explicitly or implicitly. In the former case, each 3D point is associated with local image descriptors such as SIFT [41], e.g., as part of reconstructing the scene via SfM, and 2D-3D matches are established viadescriptor matching. In the latter case, the 3D scene geometry is stored implicitly in the weights of convolutional neural networks (CNNs) [5, 7] 
%or through a random forest [6, 14, 24, 45, 67, 76]. 
The 3D-based methods are very accurate, their main drawback is that they are scene-specific, i.e., a 3D model needs to be build %or a new CNN trained 
for each new scene %. In addition, the representations need to be
and updated every time the scene changes.
%which is computationally expensive. %~\cite{[62]}.

%\paragraph{Zhou19: Learn or not to learn}
%A more flexible scene 
Representing a scene as a simple set of 2D pose-annotated images is more flexible~\cite{sattler_understanding_2019,zhou2019learn}. %with associated camera poses [62]. 
On one side, it represents all information needed to infer the 3D scene geometry; on the other side, it can be easy updated by adding more images.
%both pose-annotated or not. %posed images [62]. 
%NB:  Keep semi-supervised for latter....
%----- 2d and Absolute pose -----
%Visual localization approaches based on poses has a flexibility of  and 
2D scene representation can be easy adapted to new scenes, comparing to 3D-based methods, and allows to deploy machine learning techniques. %relative pose estimation seems a natural thing to do. 
%This leads to the question why learning-based approaches do not perform well in this setting.

%===========================================
\begin{figure}[ht]
\centering{
\includegraphics[width=0.8\textwidth]{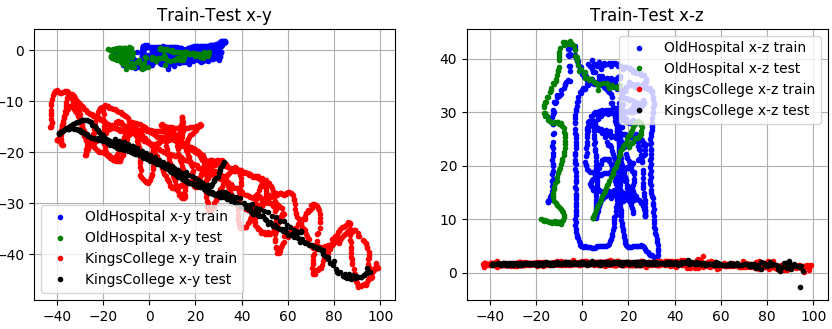}a)
\includegraphics[width=0.8\textwidth]{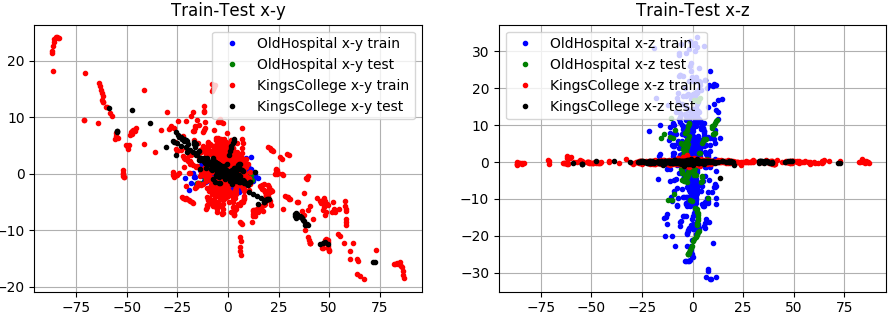}b)
}
\caption{Joint 2D projections of OldHospital and KingsCollege scenes, in position 
%(x-y, x-z, in meters)  and orientation (a-b and a-c, in degrees) 
3-DoF space. a) absolute poses; b) relative poses. Better seen in colors.}
\label{fig:domainshift-2d}
\end{figure}

%============================================

PoseNet~\cite{Kendall2015PoseNet:Relocalization} 
%and its multiple modifications
%~\cite{kendall_geometric_2017,walch17image,kendall16modelling,brachmann18learning} 
was first to cast camera localization as a regression problem.
%where camera location is directly estimated by a CNN. %pre-trained on ImageNet dataset. %classification data. %~\cite{8}. 
%Multiple improvements to PoseNet have been proposed, %to the PoseNet architecture, 
%including new loss functions %for training
%,e.g., using a weighted combination of position and orientation errors
%~\cite{kendall_geometric_2017,walch17image}, geometric re-projection errors~\cite{kendall16modelling} or adding visual odometry constraints~\cite{brachmann18learning}. 
%but all APE approaches struggle to generalize beyond their training data
%Directly regressing the absolute camera poses constrains the machine learning models to be trained and evaluated scene-wise.
%when the scenes are registered to different coordinate frames. 
The trained model learns a mapping from image to absolute pose which is dependent on the coordinate frame of the train set. % belonging to a particular scene. %This 
%causes complications, especially if one is interested in localization across several scenes simultaneously, and also 
%This prevents transferring learned knowledge of geometric relations between scenes. 
%Table~\ref{tab:naive} demonstrates the lack of generalization of learning-based methods on the Cambridge Landmarks dataset~\cite{kendall_geometric_2017}. 
%Pose estimation 
Models learned on one scene's train set work well on the same scene's test set but fail on other scenes. A typical example is shown in Figure~\ref{fig:domainshift-2d}.a) which plots poses of OldHospital and KingsCollege scenes from Cambridge Landmarks dataset in the common coordinate frame. Absolute poses of the two scenes indeed differ in range and spread\footnote{We show 3-DoF pose positions only; pose orientations show the same phenomenon.}.

%---------------- relative poses -------------------
The problem of generalization is better addressed by relative pose regression~\cite{laskar_camera_2017}. Visual localization with relative poses commonly relies on image retrieval where a query image is compared against the database of images and its pose is inferred from poses of the retrieved images.

%The latter approach assumes that images taken from the same places consists of the same landmarks and, thus would have similar feature representations. 
%These representation can be learned using full supervision to be robust to different variations in capture conditions like time of the day and weather.

%A simple way to use this representation for localization is to approximate the pose of a query image by the pose of the most similar database image [1, 72].
%identified through image retrieval [51, 68], or a combination of the poses of the most similar images [73]. 
%More accurate pose estimates can be obtained by triangulating the pose of the query image from relative poses to retrieved images [83], by using depth maps if available for absolute pose estimation [70],
%or by performing a local SfM reconstruction [62]. 

Multiple deep learning methods %based on % and regression 
have been proposed to estimate the relative pose%of the query relative to the database images
~\cite{balntas2018,laskar_camera_2017,melekhov_relative_2017,radwan2018vlocnet,saha18improved}. %[4, 35, 44, 75, 84]. % rather than to compute it explicitly from feature matches [83]. 
%However, it was recently shown that such approaches do not consistently perform better than a simple retrieval approach that only approximates the query pose [63].
%as the 
Relative poses indeed reduce the difference in range and spread but solve generalization to some extend only. Figure~\ref{fig:domainshift-2d}.b) demonstrates the phenomenon by plotting the two scenes' relative poses.
%the learned models fail to generalize on relative poses.
%Absolute pose estimation, global values live in different segments of the global 
%Generalization fails even for relative pose estimation~\cite{}
2D projections show that relative poses (\textcolor{blue}{blue} and \textcolor{green}{green} vs \textcolor{red}{red} and \textcolor{black}{black}) are still sparse. Except the space origin, they still live in different segments of the %3D position and 3D orientation 
6D pose space. %they are closer in the values but differ in spread and "shapes". $\mathbb{R}^3$
Being discriminative, %and trained on one scene, 
the supervised regression model can not generalize to poses not seen in the train set~\cite{sattler_understanding_2019}. %Moreover, the performance still remains behind the geometric methods.

State of the art methods, both 3D-based~\cite{yang2019sanet} and 2D-based~\cite{balntas2018,laskar_camera_2017,zhou2019learn} % take multiple scenes to train a relative pose model. ~\cite{laskar_camera_2017,zhou2019learn} 
test with generalization by holding out one scene in Cambridge Landmarks and 7Scenes datasets 
for evaluation and training a relative pose model on the all remaining scenes. 
% before testing one one new scene. %From the ML point of view, 
Concatenating multiple scenes allow to better populate the relative pose space and to
reduce the dataset shift when testing the model on the evaluation scene. %can be generalized from training examples. %$\mathbb{R}^6$, 

However, this reduction is rather modest. Considering relative poses as 6D points, 
only 8.8\% of relative poses in KingsCollege's test set (see Fig.~\ref{fig:domainshift-2d}.b) have a 1-top neighbour in OldHospital's train set. This fraction raises to merely 22.6\% if the train set is a concatenation of OldHospital's, StMarysChurch's and ShopFacade's sets. % train sets.
Put all together, relative poses in Cambridge Landmark dataset occupy less than 7\% of the 6D sphere with the radius equal to the average distance between two image poses. Taking the curse of dimensionality, many more annotated scenes are needed to densely populate the 6D relative pose space.

We argue that in order to progress in pose generalization one should take into account the {\it relative pose sparsity} and the {\it dataset shift} it provokes. One solution is to create a very large dataset (like ImageNet for image classification) by massively annotating scenes with poses. 
Such annotation is arduous and expensive, so the learner can deploy {\it multi-task learning} to optimize the performance across $m$ tasks/scenes simultaneously, through some shared knowledge. However, this would not solve the output shift when testing the model on a new, unseen scene.

To cope with the dataset shift in pose estimation, we %go beyond the supervised learning and 
turn towards {\it transfer learning} and develop a method 
%We address the generalization on the ..requires less posed annotated data and 
to transfer a pose model from one scene to another. 
Then we extend the method to process multiple scenes as input.

{\bf Domain adaptation.} 
The lack of generalization is a fundamental problem in machine learning. %adaptation. and addressed by the notion of dataset bias~\cite{}.
%Different conditions when collecting 
%When annotated 
Samples collected in different places and under different conditions %, it 
results in the dataset bias when a learning method trained on one dataset generalizes poorly across other datasets~\cite{torralba11unbiased}.
%In object recognition, for example, training images may be collected under specific conditions involving camera viewpoints, backgrounds, lighting conditions, and object transformations. In such situations, the classifiers learned on samples from one dataset cannot be directly applied to other related datasets.

Domain adaptation tries to produce good models on a target domain, by training on source labeled images and leveraging unlabeled target images as supplementary information during training. 
It %Domain adaptation 
has demonstrated a significant success in
% various applications, including 
image classification~\cite{HoffmanX13Efficient}, object recognition~\cite{LongCVPR14Transfer} and semantic segmentation~\cite{zhang17curriculum}.
%In this paper, 
%we propose domain adaptation as an alternative towards a better generalization of camera pose estimation. 
%To benefit from %for in the image classification and semantic segmentation. 
%We extend recent domain adaptation techniques %and extend them 
%to pose estimation, which is %not a classification but 
%a multi-regression task.
%with separate networks for learning the position and orientation of the pose.
%============================================================
%\paragraph{From classification to regression.}
These successes 
%of domain adaptation in classification and semantic segmentation is 
are due to a common semantic space shared by source and target domains. Common classes implicitly structure the output space~\cite{you2019universal}, where separation between two classes in the source can be transferred to the target. Moreover this knowledge 
%is integrated in many domain adaptation methods for classification; moreover it 
makes possible {\it unsupervised} domain adaptation~\cite{ganin2016,HoffmanX13Efficient,LongICML15Learning}.

%The domain shift includes input shift ans output shift.
%The main difficulty is in the output shift.

%Pose estimation is %not a classification but 
%a multi-regression task 
Unlike classification, domain adaptation for multi-regression tasks is less studied~\cite{borchani2015survey,lathuiliere2018comrehensive}.
Existing methods still assume the same output space and proceed by adjusting the loss function~\cite{cortes2011domain} or by reweighting source instances~\cite{chen16robust}. 
In camera pose estimation, one scene does correspond to one domain, but the difference in relative poses breaks the common space assumption.
%Lack of poses shared by source and target scenes makes an unsupervised scene adaptation questionable.
%\footnote{In the following, we use terms of scene and domain interchangeably.}.
%----------------------------------------
%Plot the paths and histograms of 
%to show the complexity of the transfer the from one domain another.
%TO DO: What would be a transformation from one scene to another, to ensure the knowledge transfer ?
%----------------------------
%Theoretical analysis assumes the alignment \cite{cortes2019domain} 
%the range an scale are solved a priori.
%Model transfer should facilitate the transfer of existing models to to new scenes or reduce the annotation effort.
%- Multi-scene the transfer can be done in supervised mode, using images without poses.
%in new conditions like in night or bad wheather condition. 
%}
%==============================================================
To extend domain adaptation to pose regression,
%We preferred the absolute pose to the relevant pose as a harder case for generalization. Also, the absolute pose methods are able to work with individual images, 
% The absolute pose regression learning and which relative pose estimation which assume working with 
%and not image sequences assumed by relative pose methods to facilitate the retrieval of image neighbours~\cite{}.
%The global pose estimation network can be main regression architecture.
%In Cambridge Landmark collection, the task is transfer a pose trained on one scene, say OldHospital, to another scene, say KingsCollege. Any pair of scene in can serve as source, target scenes.%
%We build on the recent advances in transfer learning and domain adaptation.
%Among multiple techniques developed in domain adaptation classification and semantic segmentation, 
we adopt the principle of domain-invariant representations~\cite{LongICML15Learning} and use adversarial learning to generate such representations~\cite{cao2018-eccv,cao2019learning,ganin2016}.

Due to the output shift, the source may have only some common poses with the target, %segments of the global coordinate system, 
%providing only unlabeled target images is simply not enough. 
%To adapt a source model, 
so we need supplementary information about target pose space. In this paper we discuss {\it pose-driven} %and {\it aggregation-driven} supplements. In the pose-driven 
adaptation, where the transfer task is supported with a small number of target ground truth poses. 
%In the aggregation-driven case, a global information about the target pose space, such as the 6-DoF means and variances, is provided.
%on level of 1-5\% percents of the image set. 
%where the about target scene is given by small fraction is available poses.
With such a supplement, we learn scene-invariant representation of source and target images and pivot the source model towards the target.

As an additional contribution, we enrich pose regression with {\it self-supervised learning} which proved its efficiency in other vision tasks~\cite{gidaris2018unsupervised,kolesnikov2019revisiting}. We apply random rotations~\cite{gidaris2018unsupervised} to an input image and train the network to predict these rotations. The method uses available training data to produce an additional self-supervision signal to the network and allows to learn a more accurate model.

%-------- comment CVPR'20 --------
%This submission proposes a method for camera pose estimation. It received two borderlines, and one weak reject due to 
%{\it limited novelty, lacking of generalization, insufficient experiments. The proposed method quite depends on training dataset. It needs re-training regression model for new target dataset once the scene changed. There is no comparison to either DL or geometric method.}
%Although the authors made some efforts in the rebuttal, they failed to persuade reviewers. After considering the rebuttal and further discussion/comments, the area chairs do not find sufficient cause to overturn the reviewers' recommendations. This decision has been confirmed by the AC panel.

%=================Rebuttal ==========================
%\paragraph{Multi-task versus domain adaptation.}
%In {\it multitask learning}, the learner optimizes the performance across $n$ tasks simultaneously, through some shared knowledge. In {\it domain adaptation}, the learner transfers some shared knowledge from source domains to target domains; learning of the source task is considered as irrelevant.

%To our best understanding, both concepts are applicable to the pose estimation problem. 

%Our paper explains the lack of generalization across scenes by the shift in data (scene images) and output (poses). And the low joint model errors a strong indicator that the pose regression adaptation is possible. We therefore follow the domain adaptation principle to transfer the knowledge to new scenes using the scene-invariant representations. 
%-----------------------------------
\section{Related work}
\label{sec:soa}

Traditional 
%methods for absolute pose estimation deploy image retrieval or structure-based methods. 
structure-based methods rely on SfM to associate 3D points with 2D images represented 
with their local descriptors. % (found through image processing). 
Matches between 2D points in an image and 3D points in the scene are then found by searching through the shared descriptor space. %Note that 
The descriptors can either be hand crafted (e.g., SIFT, SURF or ORB) or
learned (e.g., SuperPoint)~\cite{leng2018local,rublee11orb}. Given a set of 2D-3D matches, a n-point-pose (PnP) solver estimates candidate poses, and the best pose hypothesis is chosen using RANSAC. The estimated best pose is typically a subject of a further refinement.

PoseNet~\cite{Kendall2015PoseNet:Relocalization} modified the GoogLeNet architecture, 
%a deep convolutional network with 22 layers (six Inception modules), 
are replaced softmax layers with fully connected (FC) layers to regress the pose. 
%Directly regressing the absolute pose with end-to-end learning offered several appealing advantages. compared to the traditional methods. 
Absolute pose estimation (APE) %regression %does not require any feature engineering, it 
relies on deep network encodings that are more robust to challenging changes in the scene such as lighting conditions and viewpoint. Comparing to %structure-based methods, which require 
3D-based methods, %and heavy computations of 2D-3D matches and PnP inside a RANSAC loop,
a trained model requires less memory and has a constant inference time. 
%In addition, transfer learning enabled effective training on commonly used medium-sized datasets [27,7]. 
%However, the localization error (translation and rotation) achieved with PoseNet is an order of magnitude larger than the error attained with state-of-the-art structure-based methods.

%To reduce the localization error, To improve the performance w.r.t. the structure-based methods, 
Multiple improvements to the PoseNet %architecture %have been proposed. 
% focusing on the encoder and localizer components. 
%Rather than addressing overfitting by regularizing the dCNN model,
include new loss functions~\cite{kendall_geometric_2017},
%,e.g., using a weighted combination of position and orientation errors
adding Long-Short-Term-Memory (LSTM) layers~\cite{walch17image}, %suggested to address the problem by adding four ,
%In this architecture, each LSTM layer operates independently in one of four directions (right, left, up, down) in order to reduce the dimensionality of the image encoding.
%A second variant was also introduced as part of this work, 
%Another modification 
the geometric re-projection error~\cite{kendall16modelling}, % of the estimated pose
%This re-projection error pose loss is defined by projecting 3D points to 2D pixels with the ground-truth and estimated poses, and taking the mean of the Euclidian distances between the projected 2D points.
%Within the end-to-end learning cluster, we identify 2 main algorithmic groups: 
%They include %adding  layers~\cite{walch17image},  
%
%adding Long-Short-Term-Memory (LSTM) layers~\cite{walch17image},
%geometric re-projection errors~\cite{kendall16modelling}, 
%Instead of using just the absolute pose error, 
% suggested to include 
and additional data sources and sensor measurements~\cite{brahmbhatt18geometry}.  
Other proposals extend the Posenet-like networks with auxiliary learning~\cite{radwan2018vlocnet,valada18deep}.
%(VLocNet~\cite{valada18deep}, VLocNet++~\cite{radwan2018vlocnet}) 
They learn additional auxiliary tasks which share representations with absolute pose estimation in order to improve its learning. 
VLocNet~\cite{valada18deep} implemented the auxiliary learning approach by jointly learning absolute and relative pose estimations. Its extended version, VLocNet++\cite{radwan2018vlocnet}, added semantic segmentation as a second auxiliary task. %Specifically, the network architecture was extended to include a network dedicated to semantic segmentation, where intermediate outputs are exchanged with the absolute pose network. This addition led to a further improvement, surpassing classical structure-based methods on indoor scenes

DSAC~\cite{brachmann17dsac} and DSAC++\cite{brachmann18learning}, combine the pose estimation with structure-based methods and local learning. In such a hybrid paradigm, methods rely on geometrical constraints and utilize a structure-based pipeline where the learning is focused on local computer vision tasks, such as 2D-3D matches.

Instead of working with position and orientation losses separately,
\cite{zhou2019learn} learns the relative pose models directly from essential matrices and combines them with geometric models. It attributes the failure of pose regression in competition with geometric methods to the inaccurate feature representation in the last network layer. 
In this paper we identify the dataset shift as an additional reason of low generalization. 
In 3D-based models, generalization efforts count on 3D point clouds and their associations with the scene images. SANet~\cite{yang2019sanet} is a scene agnostic neural architecture for camera localization, where model parameters and scenes are independent from each other.
%Their method works with scene images and associated 3d point clouds;  these 3d point clouds are obtained through the dense MVS reconstruction. 
The method constructs scene and query feature pyramids, deploys the 3D point clouds at different scales and proceeds with query-scene registration (QSR). It first estimates a scene coordinate map and then computes the camera pose. %In the evaluation, for example, for the Cambridge Landmarks dataset, they train the network using 5 out 6 scenes, and test on the remaining one.

\paragraph{Domain Adaptation.}
%Domain adaptation is a well established research domain that copes with the dataset bias problem. It has witnessed %for 15 years of fruitful history, 
%a remarkable progress over last 15 years, by moving from shallow to deep methods, from an instance transfer to distribution divergence, from semi-supervised to unsupervised learning~\cite{csurka17}.

%Recent methods try to explicitly reduce domain divergence upon the deep learning framework can further exploit domain invariant features~\cite{krizhevsky12imagenet}. 

%A number of big families have been identified in domain adaptation research~\cite{csurka17,wang18deep}, including methods trying to match statistic moments on domain distributions, i.e. maximum mean discrepancy (MMD)~\cite{GrettonJMLR12Kernel} 
%and its joint and conditional variants~\cite{LongTKDE16Deep,long2017cond}. 
%\cite{tommasi_deeper_2015}.% [25, 14, 18, 17].  PanTKDE10Survey,
%or Central Moment Discrepancy~\cite{}
%(CM, and second-order statistics matching [22]. 
%\cite{long_unsupervised_2016} Unsupervised Domain Adaptation with Residual Transfer Networks.
%Another family tries 
%to generate domain-invariant representations is using the optimal transport. 
%finding  has shown excellent properties, in particular since it allows to train a unique classifier effective in all domains. 
%to align representations in source and target domains using optimal transportation~\cite{courty2017} or by associating source and target data in an embedding space~\cite{assocDA2017}.
Domain adaptation considers the discrepancy between training and testing domains as a fundamental obstacle to  generalization~\cite{LongICML15Learning,wang18deep}. The state of the art approaches address the problem by learning domain-invariant feature representations through adversarial deep learning~\cite{cao2019learning,ganin2016,you2019universal}. 
%
%Most recent methods are based on adversarial learning~\cite{cao2018-cvpr,cao2018-eccv,GaninJMLR16Domainadversarial}.
%Following the GAN principle~\cite{GoodfellowNIPS14Generative}, 
These methods encourage samples from different domains to be non-discriminative with respect to domain labels. Ganin et al.~\cite{ganin2016} 
%were first to develop a GAN-based UDA method. They 
uses a domain classifier to regularize the extracted features to be indiscriminate with respect to the different domains. They assumed the existence of a shared feature space between domains where the distribution divergence is small. The domain-adversarial neural network can be integrated into the standard deep architecture to ensure that the feature distributions over the two domains are made similar. 
%The network consists of shared feature extraction layers and two classifiers. 
%DANN minimizes %the domain confusion loss and
%label prediction loss while maximizing domain confusion loss via the gradient %reversal layer. %~\cite{GaninJMLR16Domainadversarial}.

Initially studied under the assumption of same classes in source and target  domains~\cite{ganin2016}, domain adaptation research has recently turned towards more realistic settings. The new extensions address {\it partial} domain adaptation~\cite{cao2018-eccv}, when the target domain does not include all source classes, {\it open set}~\cite{Saito_2018_ECCV} when target include new classes, and {\it universal} domain adaptation~\cite{zhou2019learn} which treat both cases jointly.

%----------------------------
%\paragraph{Partial Domain Adaptation~\cite{cao2018-eccv}} 
%In the adversarial learning family, the open set adaptation adaptation labels allow new target classes and treat them as "unknown"~\cite{Saito_2018_ECCV},  partial domain adaptation %as a new domain adaptation scenario, which 
%relaxes the fully shared label space assumption to that the source label space subsumes the target label space. 
%Previous methods typically match the whole source domain to the target domain, which are vulnerable to negative transfer for the partial domain adaptation problem due to the large mismatch between label spaces. 
%\cite{cao2018-cvpr} proposed the Selective Adversarial Network (SAN)
%Partial Adversarial Domain Adaptation (PADA), 
%in order to distinguish between the relevant and irrelevant source classes. It tries simultaneously %alleviates negative transfer 
%to down-weight the contribution of irrelevant classes when training the source classifier and domain discriminator, and to %promotes positive transfer by 
%match the feature distributions in the shared label space.

%-------- Semi-Suprevised Regression ---------------
%Semi-supervised regression: survey~\cite{kostopoulos2018ssr}

%\cite{Li2017LearningSP}:Learning Safe Prediction for Semi-Supervised Regression

%\paragraph{Target shift in adversarial domain adaptation}
%\cite{Li2019_ontarget} On target shift.

%-----------------------------------

%============================================
\section{Camera Pose Regression}
\label{ssec:pose_regression}

Given an RGB image $\bfx \in \mathbb{R}^{h\times w \times 3}$, our task is to predict the (absolute or relative) camera pose $\bfp = [\bft, \bfq]$ given by position vector $\bft \in \mathbb{R}^3$ and orientation quaternion $\bfq$, $\bfq \in \mathbb{R}^4$.
The following loss function is used to train a pose regression network~\cite{kendall_geometric_2017} 
%---------------------------
\begin{equation}
    L_{p} ({\hat \bfp}, \bfp)= ||{\bft} -{\hat \bft}|| e^{-s_t} + s_t + ||\bfq - {\hat \bfq}|| e^{-s_q} + s_q, 
\label{eq:lpose}
\end{equation}
%---------------------------
where ${\hat \bfp}=[\hat \bft,\hat \bfq]$, and $\hat \bft$ and $\hat \bfq$ represent the predicted position and orientation, respectively, $s_t$ and $s_q$ are trainable parameters to balance both distances, and $|| \cdot ||$ is the $l_1$ norm.

%{\it CNN architectures for pose regression.}
All pose regression %PoseNet-like 
networks 
%several major componentsA schematization of the PoseNet’s architecture, they all 
share three main components, namely, encoder, localizer and regressor~\cite{sattler_understanding_2019,shavit_introduction2019}.
Given an image $\bfx$, encoder $E$ is a deep network that extracts visual feature vectors from $\bfx$, $f=E(\bfx)$. The localizer then uses FC layers to map a visual feature vector to a localization feature vector. Finally, two separate connected layers are used to regress ${\hat \bft}$ and ${\hat \bfq}$, respectively, giving the estimated pose $\hat \bfp =[{\hat \bft},{\hat \bfq}]$. 

Existing variations of this architecture concern the encoder network (GoogleNet~\cite{Kendall2015PoseNet:Relocalization} and ResNet34 with global average pooling~\cite{brahmbhatt18geometry}), and the localizer (1 to 3 FC layers, 1 FC layer extended with 4 LSTMs~\cite{walch17image}, etc.). %In the following, 
We use a %most advanced PoseNet-like 
configuration that includes the ResNet34 encoder, 1 FC layer localizer and trained with the pose regression loss in Eq.\ref{eq:lpose}. 

Absolute pose estimation methods~\cite{kendall16modelling,kendall_geometric_2017,walch17image} work with the absolute poses $\bfp$. To get relative poses, we follow AnchorNet~\cite{saha18improved} and explicitly define a set of anchor points, which correspond to a subset of all training images in the network, we then estimate the pose of a test image with respect to these anchors. 
%As such, it needs to be trained explicitly per scene. 
%In contrast, our SIFT+5Pt approach does not require any training at all and can easily be adapted to new scenes.
%-----------------------------------
%==================================================
\begin{figure}[ht]
\centering{
\includegraphics[width=0.6\textwidth]{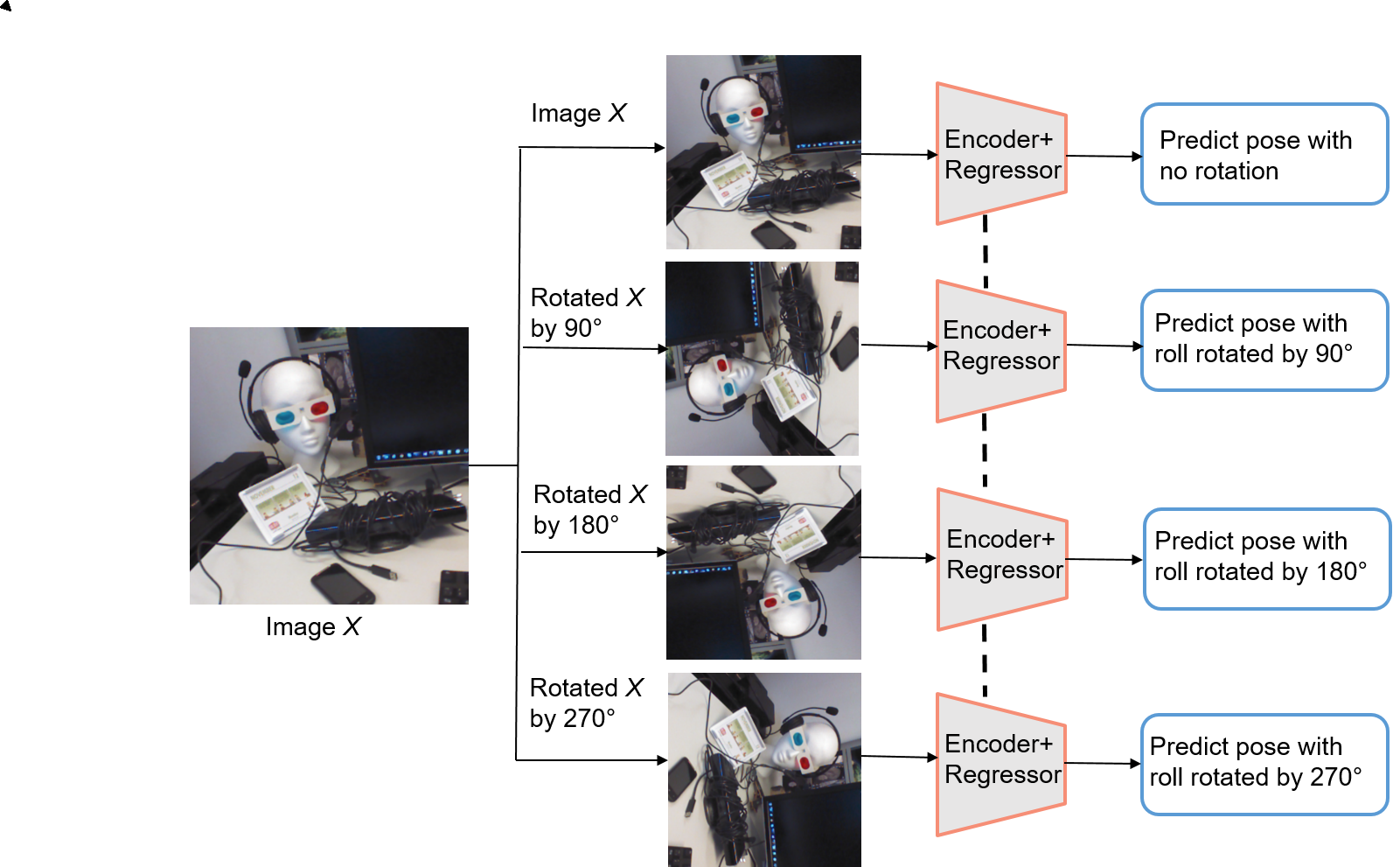}}
\caption{Self-supervised learning by rotating the input images. The model learns to predict which rotation is applied.}
\label{fig:rotation}
\end{figure}
%==================================================

%==================================================
\subsection{Self-supervision}
\label{sssec:elf}
We extend our pose estimation network with self-supervised learning~\cite{gidaris2018unsupervised,kolesnikov2019revisiting}. This concept proposes to learn image representations by training the network to recognize the geometric transformation applied to an input image. It first defines a set of discrete geometric transformations, then those geometric transformations are applied to each input image. The produced transformed images are fed to the model that is trained to recognize the transformation. 
%In this formulation, it is the set of geometric transformations that actually defines the classification pretext task that the ConvNet model has to learn. 
%Therefore, in order to achieve unsupervised semantic feature learning, it is of crucial importance to properly choose those geometric transformations (we further discuss this aspect of our methodology in section 2.2). 

We follow~\cite{gidaris2018unsupervised} in defining the geometric transformations as the image rotations by 0, 90, 180, and 270 degrees. Unlike~\cite{gidaris2018unsupervised} where the CNN model is trained on the 4-way image classification task to recognize one of the four image rotations, we train the network to identify the rotation applied to the input image (see Figure~\ref{fig:rotation}). The main argument is that in order a CNN model to be able recognize the rotation transformation that was applied to an image it will require to understand the concept of the objects present in the image, such as their location  and their pose~\cite{kolesnikov2019revisiting}.

We crop the input image %by $224 \times 224$ 
and rotate it randomly 0, 90, 180 or 270 degrees and expect that the network is able to predict the rotation applied to the image. 

Image rotation changes its orientation $\bfq$ but not position $\bft$. We calculate the orientation of the rotated image by first transforming the quaternion $\bfq$ of the input image in Euler angles %$\alpha, \beta, \gamma$, 
$yaw, pitch, roll$. 
% representing the orientation of the input image.
%Starting from the "parked on the ground with nose pointed North" orientation of the aircraft, we can apply rotations in the Z-X'-Z'' order:
%    Yaw around the aircraft's Z axis by α
%    Roll around the aircraft's X' axis by \beta
Then, we change $roll$ accordingly to the applied rotation, %by 0, 90, 180 or 270 degrees
%Yaw (again) around the aircraft's new Z'' axis by γ.
%to get the orientation of the rotated image 
%represented by the 3 Euler angles ($\alpha,\beta+rot,\gamma$) 
and transform the new %Euler 
angles back in quaternion $\bfq'$. Fed with the rotated image, we train the network to predict pose of the rotated image to be $[\bft,\bfq']$. 
 
%-----------------------------------
%------------------------------------------------
\begin{figure*}[ht]
    \centering
    \includegraphics[width=0.85\textwidth]{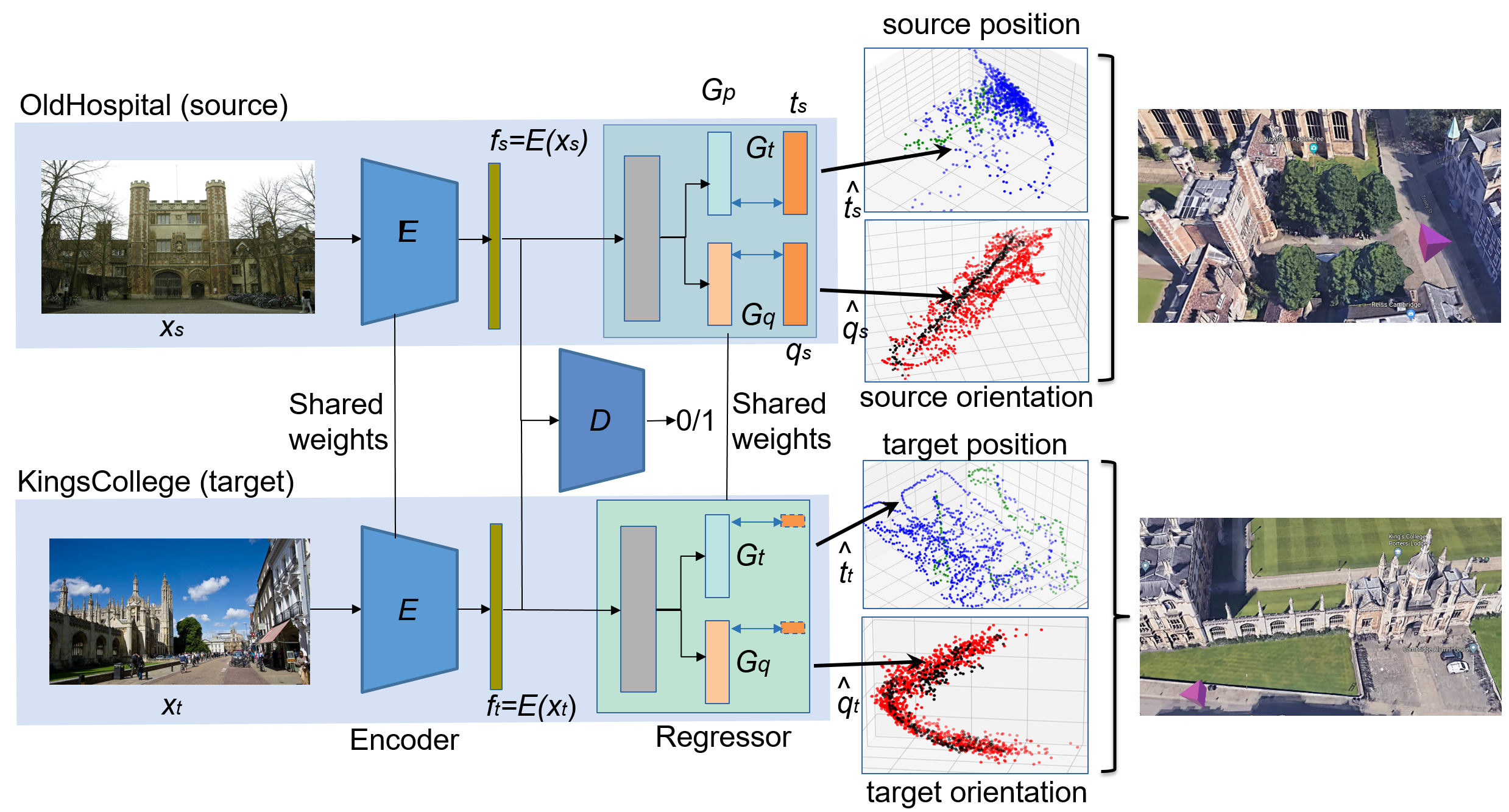}
    \caption{Adversarial Pose Adaptation Network for the camera pose adaptation, where $E$ is the transferable encoder, $G_p$ is the adaptive pose regressor (including the localizer), $D$ is scene discriminator.}
    \label{fig:arch}
\end{figure*}
%------------------------------------------------
%================================================
%---------- \cite{cao2019learning} ------------
\subsection{Adversarial Pose Adaptation Network}
\label{ssec:pose_transfer}

We now consider the task of adapting a pose regression model from one (source) scene to another (target) scene. The task constitutes a source set $D_s=\{(\bfx_s^i, \bfp_s^i)\}^{n_s}_{i=1}$ of $n_s$ images with ground truth poses and a target set $D_t=\{{\bfx_t^j}\}^{n_t}_{j=1}$ of $n_t$ images. A small number of target images might be labeled with poses, $D^a_t=\{(\bfx_t^j, \bfp_t^j)\}^{n^a_t}_{j=1}, n^a_t << n_t$.
Unlike the classification and semantic segmentation, where source and target domains share the same classes, we additionally face the output shift, where source and target poses lie in different segments of the coordinate system, i.e. 
%both their position and orientation vectors, 
%$\{\bfp_s\} \cap \{\bfp_t\} = \emptyset$ 
$\{\bfp_s\} \neq \{\bfp_t\}$. % (see Figure~\ref{fig:domainshift-2d}).
%Note that in PDA the source domain label space C s is a superspace of the target domain label space C t i.e. C s ⊃ C t . The source and target domains are drawn from different probability distributions p and q respectively. Besides p 6 = q as in standard domain adaptation, we further have p C t 6 = q in partial domain adaptation, where p C t denotes the distribution of the source domain data in label space C t .
%Output shift

Our pose adaptation network enables an end-to-end training of a transferable encoder $E$ and an adaptive pose regressor $G_p$. %Encoder $E$ is able to transfer feature extraction from source to target scene, and
Trained on labeled source images and (mostly) unlabeled target images, the network enables an accurate adaptation of the source pose model to the target scene.
%The network is trained on the source scene and makes accurate predictions on target scene.
%to sufficiently close the distribution discrepancy across scenes.
%and bound the target risk $Pr (x,y)∼q [G_y (G_f (x)) 6 = y].$

The main problem of domain adaptation is to reduce the discrepancy between the source and target images~\cite{LongICML15Learning}. Domain adversarial  networks~\cite{cao2019learning,ganin2016,you2019universal} tackle this problem by learning {\it scene-invariant} image representations in a two-player minimax game. The first player is a scene discriminator $G_d$ trained to distinguish feature representations of source images from target images, and the second player is the encoder $E$ trained simultaneously to deceive the domain discriminator $G_d$.

Specifically, the scene-invariant image representations $f=E(\bfx)$ are learned in a minimax optimization procedure, where encoder $E$ is trained by maximizing the loss of scene discriminator $G_d$, while $G_d$ is trained by minimizing its own scene discrimination loss. As the ultimate goal is to learn a source pose regression model and transfer it to target scene, the loss of the source pose regressor $G_p$ should be also minimized. 

This leads to the optimization problem over the following terms.
%proposed in~\cite{ganin2016}:
%----------------------------------
%\begin{equation}
%\begin{tabular}{cc}
%${\cal L}_{total}$ = &
%   $\frac{1}{n_s} \sum_{\bfx_i \in D_s} L_p (G_p (E (\bfx_i)),\bfp_i)$
%   \\ 
%                     & 
% $-\frac{1}{n_s+n_t} \sum_{\bfx_i \in D_s\cup D_t} L_d (G_d (E(\bfx_i)), d_i),$ 
%\end{tabular}
%\label{eq:loss_all}
%\end{equation}
%-----------------------------------
\paragraph{Source pose regression.}
The source regression loss is defined on labeled source images,
\begin{equation}
{\cal L}^s_{pose}(E,G_p)= \sum_{\bfx_i \in D_s} L_p (G_p (E (\bfx_i)),\bfp_i),
\label{eq:loss_src}
\end{equation}
%-----------------------------------
where $L_p$ is the regression loss function defined in Eq.~\ref{eq:lpose}.

%-------------------------------------
\paragraph{Scene discrimination network.}
Scene discriminator $G_d$ is trained to distinguish between feature representations of the source and target images, with the adversarial loss
%----------------------------------
\begin{equation}
{\cal L}_{adv}(G_d)= -\sum_{\bfx_i \in D_s\cup D_t} L_d (G_d (E(\bfx_i)), \bfd_i),
\label{eq:loss_adv}
\end{equation}
%-----------------------------------
where $L_d$ is the cross-entropy loss function and $\bfd_i$ is the scene label (0 for source images and 1 for target images).

%=================================================
\paragraph{Semi-supervised adaptation.}
In classification, domain adaptation can be achieved with the two above terms in an unsupervised way~\cite{ganin2016,you2019universal}. In pose regression, %scenes have no common pose space. According to the scene-driven scenario, 
we are supplied with a small number of labeled target images, $D_t^a$. Then we define a regression term on $D_t^a$, 
%-----------------------------------
\begin{equation}
{\cal L}^t_{pose}(E,G_p)= \sum_{\bfx_j \in D^a_t} L_p (G_p (E (\bfx_t^j)),\bfp_t^j),
\label{eq:loss_trg}
\end{equation}
%\frac{1}{n^a_t} 
%-----------------------------------
where $L_p$ is the regression loss function defined in Eq.~\ref{eq:lpose}. The pose regression loss then includes the source and target terms, ${\cal L}_{pose}(E,G_p) = {\cal L}^s_{pose}(E,G_p) +{\cal L}^t_{pose}(E,G_p)$.

%In the image with available pose participate in two losses, to reduce the regression loss and as target image to reduce the adversarial loss

%One scene regression loss in Equation~\cite{} 
%-------- \cite{lou2019CloserLook}
The total loss for training our {\it adversarial pose adaptation network} (APANet) can be represented as
%--------------
\begin{equation}
    {\cal L}_{APANet}(E,G_p,G_d) = {\cal L}_{pose}(E,G_p) +\alpha {\cal L}_{adv}(E,G_d),
\label{eq:loss}    
\end{equation}
%--------------
where $\alpha$ is a hyper-parameter controlling the importance of the adversarial loss.

The training objective of the minimax game is the following
%------------------------------
\begin{equation}
E^{*}, G_p^{*} = \arg \min\limits_{E,G_p} \max\limits_{G_d} {\cal L}_{APANet} (E,G_d,G_p).
\label{eq:minmax}
\end{equation}
%-----------------------------
Eq.~\ref{eq:minmax} is solved by alternating between optimizing $E,G_p$ and $G_d$ until the total loss (\ref{eq:loss}) converges.

The APANet architecture is presented in Figure~\ref{fig:arch}. The network inputs a batch of source images and a batch of target images. Encoder $E$ generates
image representations $f=E(\bfx)$ for both batches. Scene discriminator $G_d$ is trained on image representations $f$ and scene labels $\bfd$ to distinguish source images from target images. Pose regressor $G_p$ is trained on a full set of source image poses and, when available, a small number of target scene poses. Position regressor $G_t$ and orientation regressor $G_q$ are trained separately. The position and orientation predictions are concatenated to produce the 6-DoF pose estimation, $\hat \bfp =[{\hat \bft},{\hat \bfq}]$, where
${\hat \bft} =G_t(E(\bfx))$ and ${\hat \bfq} =G_q(E(\bfx))$.

\subsection{Scene adaptability}
\label{ssec:adaptability}
%Error of ideal joint hypothesis. 
We complete this section by the notion of adaptability which has been introduced~\cite{chen19transferability} to measure the transferability of feature representations from one domain to another. %It ass the ideal joint hypothesis $h^∗$
Adaptability is quantified by the error of a joint hypothesis $h^{*}$ in both domains.
%as $\lambda = err^s (h^∗) + err^t (h^∗)$. 
%
%ResNet-50 (He et al., 2016) pre-trained on ImageNet (Russakovsky et al., 2015), Domain Adversarial Neuralhttps://www.overleaf.com/project/5e4e4da9dff2ac0001427aa1 Network (DANN)~\cite{ganin2016}, and Maximum Classifier Discrepancy (MCD) (Saito et al., 2018). 
The ideal joint hypothesis $h^{*}$ is found by training on both source and target labeled images. Note that the target labels are only used to reason about the adaptability. 
%The error of the ideal joint hypothesis on the source domain, the target domain, and their sum $\lambda$ are shown in Figure 2(b). It is somewhat unexpected that the adaptability $\lambda$, as quantified by the error of the ideal joint hypothesis $h^∗$, worsens substantially in the adversarial feature adaptation methods DANN and MCD, compared to the non-adaptation method ResNet-50.
%We compute it by training a joint regression model over the feature representations using the baseline method (see Section~\ref{ssec:pose_regression}).

If the joint model shows a low error on source and target test sets, 
%(comparable to the individual domain errors), 
it suggests that an efficient transfer is possible across domains. In the evaluation section, we apply this idea to pose adaptation tasks. We will learn a joint model in order to reason about adaptability of source absolute and relative models to the target scene.

%-----------------------------------
\section{Evaluation}
\label{sec:evaluation}

%---- \cite{Li2019_relative} -----
%We test our APANet on two publicly available camera relocalization benchmark datasets. % one indoor and one outdoor, %to demonstrate its effectiveness. 
%Experimental results are presented and compared with state-of-the-art methods in the literature.
%We also investigate the role of network components and analyze how different choices affect the network performance. 
%==========================================
%\vspace{-2mm}
\paragraph{Datasets.}
%\label{ssec:dataset}
We test our pose adaptation network on two public datasets, Cambridge Landmarks~\cite{Kendall2015PoseNet:Relocalization} and 7Scene~\cite{shotton13scene}.
%We use the same split of train and test set split proposed in the original datasets.
%
{\it Cambridge Landmarks}~\cite{Kendall2015PoseNet:Relocalization} is an outdoor dataset collected in four sites around Cambridge University. It is collected using a Google mobile phone while pedestrians walk. The images are captured at the resolution of 1920x1080, the ground truth pose is obtained through VisualSFM software. Each site corresponds to one scene: Old Hospital, King's College, StMary's Church and Shop Facade. 
%The dataset is also very challenging as it is taken in different weather and lighting conditions. Besides, the occlusion of moving pedestrians and vehicles further increases the difficulty. By analogy with the domain adaptation protocol~\cite{cao2018-eccv}, 
We first consider 1-to-1 scenario %We use the 4 scenes to 
to test the model transfer from one scene to another. We form twelve pose adaptation tasks, by enumerating all possible {\it  source} $\ra$ {\it target} pairs. 
Then, we consider $n$-to-1 scenario and form four adaptation tasks, where 
one scene is retained as target and three remaining scenes is used as source.
%3 scenes are concatenated to form the source and the remaining scene is used as target.

{\it 7Scene}~\cite{shotton13scene} is an indoor dataset 
%for camera relocalization and trajectory tracking. It is 
collected with a handheld RGB-D camera. The ground truth pose is generated using the Kinect Fusion approach~\cite{shotton13scene}. The dataset is captured in 7 indoor scenes. 
%For each scene, it contains several image sequences, which has already been divided into training and testing sets.
In 1-to-1, we form twelve adaptation tasks, with Chess selected as a pivot scene. It constitutes six Chess $\ra X$ tasks and six $X \ra$ Chess tasks, where $X$ is one of the six remaining scenes: Fire, Heads, Office, Pumpkin, Red Kitchen or Stairs.
%The images are taken at the resolution of 640x480 with known focal length of 585. The dataset is quite challenging as motion makes the images blur. Besides, the indoor scenes are usually texture-less, which makes the localization problem even more difficult.
In $n$-to-1 scenario, seven adaptation tasks are formed, 
one scene is retained as target and six other scenes is used as source.
%where 6 scenes are concatenated to obtain the source and the remaining scene is used as target.

%and their learning rates are 10 times that of the fine-tuned layers. We use mini-batch SGD with momentum of 0.9 and the learning rate decay strategy implemented in DANN [10]: the learning
%$\nu_0$ rate is adjusted during SGD using $\nu_p =\frac{\nu_p} {(1+\alpha p)^\beta}$, where $p$ is the training progress linearly changing from 0 to 1. The
%flip-coefficient of the gradient reversal layer is increased %gradually from 0 to 1 as DANN [10]. Hyper-parameters are optimized with importance weighted cross-validation [35].

%----------------------------------------
\begin{table}[ht]
\begin{center}
\scalebox{0.85}{
\begin{tabular}{|l|l|l||r|r|r|} \hline
Source &Case &  Method  & OldHospital & StMarysChurch& ShopFacade   \\ \hline
%Kings    &3D &Active Search    &\bf{0.44/1.01} & \bf{0.19/0.54}  & {\bf 0.12/0.40}\\ 
Kings     &3D &DSAC++~\cite{brachmann18learning} &\bf{0.20/0.30} & \bf{0.13/0.40}  & {\bf 0.06/0.30}\\ 
\cline{2-6} % removed  KingsCollege {\bf 1.04/4.76}
College   &APE &No adaptation    &38.63/83.04  & 28.53/110.39 & 31.21/45.19 \\ 
%         &&Transposed    &12.77/15.31 & 17.09/44.02 & 8.17/21.62 \\ \cline{3-6}
          &&Joint          &       1.55/5.05   &   2.19/6.65  & 1.32/4.58 \\
\cline{3-6}
          &&SS, $\nu$=0.05  &      6.74/14.92  &   6.76/15.84 &  6.52/13.10 \\
          &&APANet,$\nu$=0.05&     3.74/8.27   &   3.63/10.18 &  2.61/8.63 \\ 
          &&APANetS,$\nu$=0.05&    3.56/6.71   &   3.58/8.23  &  2.58/7.41 \\
\cline{2-6}  
%          & AnchorNet        &     1.21/2.55  &    1.04/2.69  &   0.52/2.27 \\ %KingColleges 57/0.880.
          &RPE &No adaptation &     3.63/7.22  &    3.53/10.39 &   2.91/8.16 \\
% +Kings  Anchornet                 0.57/0.88       0.57/0.88      0.57/0.88
          &    &Joint         &     0.53/1.59  &    0.45/1.21  &   0.42/1.29 \\ \cline{3-6}
          &&SS,  $\nu$=0.05   &     1.54/1.82  &    1.36/3.04  &   1.18/3.81 \\
          &&APANet,$\nu$=0.05 &     1.03/2.17  &    0.80/2.18  &   0.61/2.62 \\   
          &&APANetS,$\nu$=0.05&     0.98/1.94  &    0.77/2.25  &   0.62/2.49 \\ \cline{3-6}
&&NC-Essnet*~\cite{zhou2019learn}&  0.95/2.65  &    1.12/3.64  &   0.70/3.41 \\ %\cline{2-6}
%&&Sift+5pt~\cite{zhou2019learn}&     0.88/1.91   &    0.35/1.58  &   0.17/0.99 \\ 
\hline
\end{tabular}
}
\caption{KingsCollege$\ra X$ adaptation tasks in Cambridge Landmarks dataset. The median position (in meters) / orientation (in degrees) errors are reported.}
\end{center}
\label{tab:cambridge}
\end{table}

%Source Target 0.9  0.64 1.47 SS, as joint
%Source Target 0.9  0.61 1.45
%Source Target 0.9  0.60 1.41

%Source Target 0.05 1.52 5.24 Average SS for nu=0.05, over 12 RPE tasks
%Source Target 0.05 1.11 2.98 Average APANet
%Source Target 0.05 1.09 2.61 Average APANEtS 
%----------------------------------------
%==========================================
\paragraph{Implementation and Setup.} 
%--- \cite{cao2019learning} -----
The APANet is implemented in PyTorch. Encoder $E$ is fine-tuned on 
%ResNet-34 [15] and  ResNet-34 [34]
ResNet-34 network. Pose regressor $G_p$ and scene discriminator $G_d$ are trained from scratch. $G_p$ includes a FC localizer with 1024 nodes and two separate layers $G_t$ and $G_q$, to regress position and orientation vectors, with 256 nodes each. Scene discriminator $G_d$ is similar to one used in the universal domain adaptation (UDA) network~\cite{you2019universal}; it includes three FC layers with 1024, 256 and 64 nodes, interleaved with ReLu layers, the drop-out is 0.5. 

In the train phase, %all APANet components are involved.
the network inputs a batch of source images and a batch of target images 
%and outputs the corresponding global poses of them. 
to fine-tune the encoder $E$ and to train scene discriminator $G_d$ and pose regressor $G_p$. 
%It is important to note that, the twin networks are identical. One takes the current image as input and produces its global 6DOF pose information, while the other takes the reference image as input and outputs its corresponding pose.
%
In the test phase, only encoder $E$ and pose regressor $G_p$ are necessary. 
%Since they are identical, any one can be used.
%The scene discriminator $G_d$ is no longer necessary in this stage
%The middle part that linking the twins . 
A target image $\bfx_t$ is fed to the network and its pose is estimated as 
$G_p(E(\bfx_t))$.

%The modified ResNet50 is initialized with pre-trained
%weights of ImageNet dataset. The GPRU component and
%the RPRU are initialized with the Xavier initialization [54].

%The scene discriminator $D$ is similar to one used in the universal domain adaptation (UDA) network~\cite{you2019universal}. It is composed of three FC layers with 1024, 256 and 64 nodes.
%
%It inputs feature vectors of source and target images provided by the . It learns to distinguish between source and target scene images, it is a scene classifier. During the training phase it is designed to distinguish between source or target scene images. $G_d$ is trained on 0/1 scene indicator values $d_i$, where 0 and 1 are associated with the source and target images, respectively.

We use the Adam optimizer; % to train the network.
%with parameters $\beta_1 = 0.9$ and $\beta_2 = 0.999$. The weight decay is $10^{-5}$.
we train the network with a learning rate of $10^{-5}$ and the batch size of 16 images. We initialize parameters $s_t$ and $s_q$ in the pose loss (Eq.~\ref{eq:lpose}) with 0 and -1.0 respectively. Hyper-parameter $\alpha$ is set to 1.0.
We use the same image pre-processing steps as the state of the art  methods~\cite{kendall_geometric_2017,sattler_understanding_2019}. 
%In Cambridge Landmarks dataset, we resize the image to 256 pixels along the shorter side and normalize it with the mean and standard deviation computed from the ImageNet dataset. 
For training, we randomly crop the image to $224 \times 224$ pixels. For testing, images are cropped to $224 \times 224$ pixels at the center of the image. Training images are shuffled before they are fed to the network.
In the self-supervised extension (APANetS), the input image is randomly rotated 0, 90, 180, 270 degrees and the network is trained to correctly predict the applied rotation. 
%\vspace{-2mm}
%===========================================
\paragraph{Evaluation options.}
%\label{ssec:options}
%Table I in the introduction section showed a multi-fold error increase in baseline schema when the rose regression
%The ultimate goal of the joint learning is to find such an invariant representation without poses, in unsupervised setting.
%For all we adopt the train/test split used in the literature.
%All training images from a scene are used to train the scene classifier $G_d$. are randomly sampled to for classification
For all adaptation tasks, we compare five models, including two baselines and two APANet versions.
%and compare and implement and evaluate  following methods.
%\begin{enumerate}
%\item 
First, we train the {\it joint model} (see Section~\ref{ssec:adaptability}) on source and target train sets. This model is an indicator of adaptability of the source model to target scene. The low error on source and target test sets suggests a good generalization and there exists a joint model performing well on both scenes.
%Joint model  represents an optimal scenario and gives an upper bound of target error.
%\item 
Second, as the true baseline, we consider a {\it semi-supervised (SS) relaxation} of the joint model, where a fraction of target train set is used for training. Denoted $\nu$, where $\nu=\frac{n_t^a}{n_t}$, this fraction varies between 0 (source scene only) and 1 (the joint model). 
%\item 

The third model is the APANet (Section 3.2) trained with $\nu$ target poses. This model can be considered as the SS baseline extended with adversarial learning of scene-invariant image representations. Finally, we include results of APANet extended with {\it self-supervision} (Section 3.1) and denoted APANetS. 
%where the target pose means and variances are used to transpose unlabeled target images into the source pose space (see Section 3.3), before training the APANet in the unsupervised mode.
%Finally, we also include results of APANet with {\it scene transposition}, where the target pose means and variances are used to transpose unlabeled target images into the source pose space (see Section 3.3), before training the APANet in the unsupervised mode.

%Instead, we assume that a tiny fraction of target images are available with their poses, on the level of 1-5\% of the scene.

%It is important to note that 
The joint model gives an estimate of the smallest error the optimal joint model could achieve on the target scene. It serves as an indicator for the SS baseline and APANets. All three methods are exposed to the trade-off between ratio $\nu$ and accuracy drop with respect to the joint model. In the following, we compare the SS baseline and APANets for the same values of $\nu$, in order to measure the size of target supplement sufficient to ensure an efficient pose mode transfer.

%DANN can work and use the annotated available to guide the transformation between the two scenes.

%===========================
%Geometric methods need no adaptation, they use the scene information in the form of  3d map, and 2d-3d correspondences to cole availablity of 3d, is replaced

%===========================================
% Manual copying, old version 
%\input{cambridge_evaluation_old.tex}

%========================================
\subsection{Evaluation Results}
\label{ssec:results}
%We compare the results of four proposed methods
%with that of state-of-the-art deep learning-based methods such as PoseNet, Bayesian PoseNet, PoseNet2, Hourgrlass-net, LSTM-Net and RelNet on the 7Scene dataset, and with PoseNet, Bayesian PoseNet, PoseNet2 and LSTM-Net 
%on the Cambridge Landmarks and 7Scene datasets. 
%Similar to others, we report each scene’s median error. We also compare the average median accuracy over all scenes in each dataset. 

%=======================================
\begin{figure}[t]
\centering{
\includegraphics[width=0.9\textwidth]{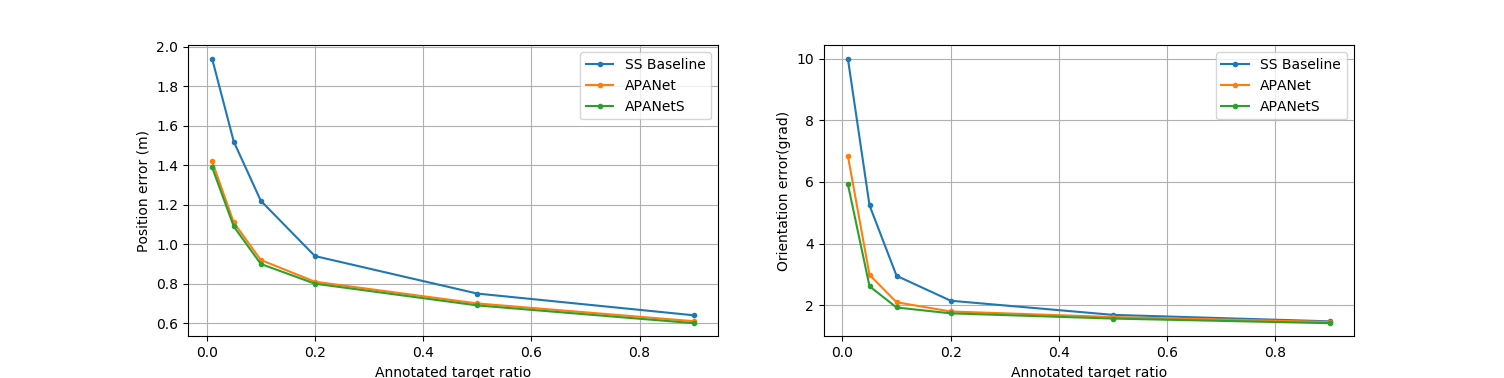} 
\caption{Position (top) and orientation (bottom) RPE errors of SS, APANet and APANetS for different $\nu$ values.}}
\label{fig:semisup}
\end{figure}
%=======================================
%=======================================
\paragraph{Cambridge Landmarks.} 
Table 1 %~\ref{tab:cambridge}
reports APE and RPE evaluation results for three adaptation tasks KingsCollege $\ra X$. 
%defined on the Cambridge Landmarks dataset. 
%It can be seen that our method obtains the best positional accuracy on the KingsCollege and the ShopFacade scenes, reaching accuracies of 0.865m and 0.834m respectively. We improve the state-of-the-art orientation accuracy of the OldHospital and the StMarysChurch scenes from \ang{3.29} and \ang{3.32} to \ang{2.42} and \ang{2.98}, achieving 26\% and 10\% improvement respectively. The average positional accuracy over all scenes is improved from 1.30m to 1.24m. The average orientation accuracy over all scenes is only a little worse than that of PoseNet2, which is trained with 3D model constraints.
For each task, the table reports the error without adaptation, the joint model error, SS and APANets errors for the selected value $\nu=0.05$, which corresponds to 5\% of target ground truth poses available for training. In all tasks, the joint model error is small 
and suggests a good adaptability across scenes. 
%comparable to the individual scene model errors (cmp. Table 1). 
%The joint position errors vary from 1.32m for KingsCollege $\ra$ ShopFacade to 2.81m for OldHospital $\ra$ ShopFacade adaptation. The orientation error varies between~\ang{4.58} for KingsCollege $\ra$ ShopFacade and~\ang{7.80} for StMarysChurch $\ra$ OldHospital adaptation. %According to the adaptability notion~\cite{zhou2019learn}, 
%This suggests a good adaptability across scenes. 
%Joint position and orientation errors 

%show an advantage is when the Smaller nu values , larger (seen next section).
%Then, we pay attention to the APTN errors an its comparative increase with respect to the joint model. 

%It is interesting to note that of all the methods presented in the two tables, some did better in positional accuracy and some did better in orientation accuracy, none of them seems to comprehensively beat the others in both measures. Our method achieves the best average positional accuracy amongst all methods in both datasets. For orientation accuracy, our method achieves competent results, which is only slightly worse than the best method (PoseNet2) but better or at least as good as the other methods.

For the ratio $\nu=0.05$, APANets perform much better that the SS baseline. Averaged over 12 adaptation tasks, the APANet position error is
46.8\% (APE) and 52.2\% (RPE) lower than the SS position error. Similarly, the APANet orientation error is on average 37.9\% and 42.1\% lower than the SS orientation error. Self-supervised image rotation in APAnetS further reduces the orientation error but has a negligible impact on the position error.
%===========================================
\begin{table*}[ht]
\begin{center}
\scalebox{0.85}{
\begin{tabular}{|l|l|l||r|r|r|r|r|r|} \hline
Source& Case & Method & Fire & Heads & Office & Pumpkin & Kitchen & Stairs  \\ 
\hline
Chess 
%  &3D&Active Search& \bf{0.05/0.32}& \bf{0.03/0.14} & \bf{  0.04/0.12}& \bf{0.05/20.82}&\bf{ 1.59/23.14}&\bf{0.83/8.45} \\ \cline{2-9}
  &3D&DSAC++~\cite{brachmann18learning} &\bf{0.02/0.90}&\bf{0.01/0.80}&\bf{0.03/0.70}&\bf{0.04/1.10}&\bf{0.04/1.10}&\bf{0.09/2.60} \\ \cline{2-9}
  
  &APE&No Adaptation   & 2.25/28.32 & 1.91/31.76 & 0.30/23.75 & 1.08/20.82 & 1.59/23.14 & 1.83/28.45 \\ \cline{3-9} % Ex transposed 
  & & Joint            & 0.40/9.45  & 0.28/14.41 & 0.31/7.20  & 0.27/4.91  & 0.34/7.26  & 0.34/9.52 \\ \cline{3-9}
  & &SS,$\nu$=0.05     & 1.31/19.08 & 1.04/24.62 & 0.89/18.36 & 0.82/11.15 & 1.01/17.59 & 1.03/20.12 \\
  & &APANet,$\nu$=0.05 & 0.63/14.67 & 0.44/17.03 & 0.42/10.62 & 0.39/6.87  & 0.43/10.35 & 0.47/12.28 \\ 
  & &APANetS,$\nu$=0.05& 0.59/12.06 & 0.41/14.29 & 0.41/8.45  & 0.37/5.71  & 0.41/8.66  & 0.48/9.87 \\ 
  \cline{2-9}

  &RPE&No Adaptaption  & 2.25/28.32 & 1.91/31.76 & 1.63/23.75 & 2.78/20.82 & 1.89/23.14 & 1.83/28.45 \\ \cline{3-9}
%  & &AnchorNet         & 0.15/10.30 & 0.08/10.90 & 0.09/5.15  & 0.10/2.97  & 0.08/4.68  & 0.10/9.26 \\ 
  \cline{3-9} % joint with kings 0.06/3.89
    
  & & Joint            & 0.11/7.14  & 0.08/7.14  & 0.09/4.02  & 0.09/4.68  & 0.07/3.86 & 0.09/8.52 \\ \cline{3-9}
  & &SS,  $\nu$=0.05   & 0.41/17.08 & 0.34/14.62 & 0.29/14.36 & 0.42/11.15 & 0.31/7.75 & 0.29/19.12 \\
  & &APANet,$\nu$=0.05 & 0.22/10.15 & 0.17/10.26 & 0.14/7.36  & 0.19/6.46  & 0.19/5.65 & 0.17/12.28 \\ 
  & &APANetS,$\nu$=0.05& 0.21/9.72  & 0.15/9.35  & 0.15/6.69  & 0.19/5.87  & 0.16/5.13 & 0.16/11.77 \\ \cline{3-9}
  &&NC-EssNet*~\cite{zhou2019learn} 
                       & 0.26/9.64  & 0.14/10.66 & 0.20/6.68  & 0.22/5.72 & 0.22/6.31 & 0.31/17.88 \\ %\cline{3-9}
%  &&Sift-5Pt~\cite{zhou2019learn} 
%                       & 0.05/2.85  & 0.04/3.86  & 0.06/2.66  & 0.08/3.81 & 0.07/3.12 & 0.22/11.08 \\
\hline
\end{tabular}}
\caption{1-to-1 transfer in 7Scenes dataset. The median position (in meters) / orientation (in degrees) errors are reported.}
\end{center}
\label{tab:7scene}
\end{table*}

%Chess APAN - 0.18m 5.17◦ 

%===========================================
\vspace{-2mm}
%===========================================
\paragraph{Impact of ratio $\nu$.}
\label{ssec:aggregation}
%We test the role of the hyper-parameter $\alpha$, fixed to 10 in Tables~\ref{tab:cambridge} and \ref{tab:7scene}.
%Note that the semi-supervised baseline is modeled by setting the hyper-parameter $\alpha$ to 0 in Eq.\ref{eq:loss}.
Figure~\ref{fig:semisup} gives an aggregated view by averaging results over twelve adaptation tasks. It compares the SS, APANet ans APANetS errors for $\nu$ varying between 0.01 and 0.90. 
%Additionally, the hyper-parameter $\alpha$ in the APANet loss (\ref{eq:loss}) is tested with values 1, 10 or 100 (in Tables 2 and 3, $\alpha$ is set to 10).

The figure shows little or no difference when target poses are abundantly available in training. Having 20\% or more target poses is sufficient to train a joint model working well on both scenes. Instead, when the ratio $\nu$ is reduced to 1\%-5\%, the advantage of APANet over the SS baseline becomes multi-fold. 
%Making image representations scene-invariant helps keep the target error low with a tiny target annotation.
%In addition, we compare the performance of both methods to the (optimal) joint model. For the SS baseline, 2-fold increase of position and orientation error is at 11.12\% and 13.56\% of ratio $\nu$, respectively. Instead, for APAN, 2-fold error increase %of the position and orientation error is at 3.37\% and 4.52\%, respectively. 
We conclude that learning scene-invariant image representations accelerates the adaptation across scenes. 
%The evaluation shows that adding the adversarial domain adaptation accelerates the
Moreover it enables an accurate adaptation with a smaller target supplement. % with respect to the baseline semi-supervised setup.

Note no existing RPE method can perform 1-to-1 scenario without performance drop.
Results for 3D-based DSAC++~\cite{brachmann18learning} and 2D-based NC-EssNet~\cite{zhou2019learn} are included in Table 1 (and Table 2) for the comparison only. The state of the art NC-EssNet is a supervised learning method trained and tested on the target scene, while APAnet(S) are transfer models trained on the source train set and a fraction of the target train set. Still, all 2D-based methods are less accurate than the best 3D-based models, such as DSAC++ and Active Search.

\vspace{-2mm}
%======================================================
\paragraph{7Scenes.} 
Table 2 %~\ref{tab:7scene}
reports evaluation results for 1-to-1 adaptation tasks Chess $\ra X$ in the 7Scene dataset.
%For each task, the table reports the transposed model error, the joint model error, and the SS and APANet errors for the selected value $\nu=0.05$. 
In all tasks, we again observe a small error of the joint models. For the ratio $\nu=0.05$, the APANet performs better that the SS baseline and 
APANetS improves over APANet in the orientation error. % better that the SS baseline and 
%Averaged over all 12 adaptation tasks, the APANet position error is 53.4\% lower than the SS position error. Similarly, the APANet average orientation error is 35.5\% lower that the SS orientation error. Note close error values for joint models of the symmetric scene pairs.

%No common protocol
%none method can generalize 1-to-1 scene accurately. In the best of cases, all (but one) scene are use to train a regression model

%\footnote{Complete evaluation and comparison for 7Scenes dataset is provided in the Appendix.}.
%It is seen that compared with 7 state-of-the-art deep learning-based camera relocalization methods, the proposed method achieves the best performance on positional accuracy in all 7 scenes. Our method improves the average median positional accuracy by 16\% over the best reported result. It is interesting to note that our method has obtained even better result than PoseNet2, which utilizes 3D reference as additional constraints.

%------------------------------------------------
\begin{table}[ht]
\begin{center}
\scalebox{0.85}{
\begin{tabular}{|l|l|l|r|r|r|r|r|} \hline

Source&Target&Case& Joint   & SS,$\nu$=0.05& Apanet,$\nu$=0.05 & APANetS,$\nu$=0.05 & NC-EssNet*\\ \hline
Cambridge&Cambridge& 1-to-1 & 0.67/1.45 & 1.90/4.08 & 0.88/3.07 & 0.87/2.86 & \\   
         &         & n-to-1 & 0.63/1.38 & 1.51/3.82 & 0.81/2.96 & 0.82/2.56 & 0.85/2.82 \\ \hline
Cambridge&7Scenes& 1-to-1 & 0.36/6.37 & 1.13/20.10 & 0.62/13.16 & 0.61/12.43 & \\ 
         &       & n-to-1 & 0.31/6.49 & 1.19/19.01 & 0.50/13.47 & 0.47/12.96 & 0.48/32.97 \\ \hline
7Scenes&7Scenes& 1-to-1 & 0.13/3.84 & 0.38/10.22 & 0.25/7.63  & 0.24/7.16 & \\
       &       & n-to-1 & 0.13/3.45 & 0.34/9.30  & 0.22/7.76  & 0.22/7.31 & 0.21/7.50 \\ \hline
7Scenes&Cambridge& 1-to-1 & 1.78/7.42 & 5.86/27.17 & 3.95/16.67 & 3.91/16.15 & \\  
       &         & n-to-1 & 1.61/7.55 & 5.81/26.08 & 3.86/15.86 & 3.87/15.56 & 7.98/24.35 \\ \hline
\end{tabular}}

\caption{RPE transfer in 1-to-1 and n-to-1 scenarios. The median position (in meters) / orientation (in degrees) errors are reported.}
\end{center}
\label{tab:aggregate}
\end{table}

%Source Target 0.05 1.52 5.24 Average SS for nu=0.05, over 12 RPE tasks
%Source Target 0.05 1.11 2.98 Average APANet
%Source Target 0.05 1.09 2.61 Average APANEtS
%------------------------------------------------
%================================================
\paragraph{Cross dataset transfer.}
Table 3 %~\ref{tab:aggregate}
reports evaluation results for 1-to-1 and $n$-to-1 scenarios, on the same or accross the Cambridge and 7Scenes datasets. %Since existing RPE methods target n-to-1 scenario 
%For the sake of comparison, 
When we test the APANets in $n$-to-1 scenario on the same dataset, a model is trained with all but one scenes, such a setup is used in the previous methods~\cite{laskar_camera_2017,zhou2019learn}.
In evaluations across the two datasets, all scenes from the source dataset are used for training.  
All the joint models behave well and indicate a good generalization. 
The self-supervised learning in APANetS reduces the orientation error. Instead, adding more scenes does really help APANet and APANetS.

%Multi-source transfer should be used instead.
%not adapted to the multiple sources, despite a good join results

Again, results of NC-EssNet~\cite{zhou2019learn} are included for the comparison only. Being totally supervised, NC-EssNet shows good performance on the same dataset (Cambridge to Cambridge and 7Scenes to 7Scenes) but suffers from the dataset shift in the cross-dataset evaluations, if  compared with APANets.

%----------------------------
%Our 'dream' i to make the model transfer totally unsupervised or model-driven (a few prior parameters to guide the transfer)

%============================================================
\subsection{Discussion}
\label{ssec:discussion}
%A low generalization of absolute pose regression is due the scene-specific knowledge implicitly encoded in CNNs trained on scene pose-annotated images. In this paper, we explored an alternative scenario, aimed to adaptation from one scene (or multiple scenes) to a new scene in unsupervised manner.  
%We benefit from the recent advances in the domain adaptation which proved their efficiency in image classification and semantic segmentation tasks.
%We show how the principle of domain invariant representations can be extended to the multi-regression tasks. %in the straightforward way.
%We promoted the data-driven approach where target images without pose annotations are easy to collect.
%Using the adaptation scenario of pose invariant image representations can be done in the
%adaptation of pose regression %from non annotated images represent a strong alternative to the standard scenario of totally unseen scenes.
%Pose regression adaptation from scene to scene represents a to the generalization for global pose estimation

We identify the dataset shift an important barrier to the generalization of pose estimation models. Analysis of the evaluation results in the previous section raises two critical issues. One concerns the target supplement to accompany the adaptation process and to preserve the pose estimation accuracy. We have promoted the pose-driven adaptation, where a small number of target poses guide adaptation with a modest performance drop. This method showed promising results. However, acquiring target ground truth poses is not always possible. 
%On the other hand, both aggregation-driven techniques discussed in Section 3 fail to align source and target scenes and result in a severe performance drop.
We therefore need alternatives to parametrize the model transfer across scenes, where a model-driven regression can alleviate the absence of target poses.
% -with no need of target poses, 
%- unsupervised model-driven transfer.

%Alternative 2 . Reduce annottaion need: under 5\% reducting the annotation need 
%still unsupervised transfer, as collecting non-annotated image in has a low cost.  

%Alternative 3. Replace one-source with a multi-source GAN adaptation 

%--- 2 \paragraph{Low adaptability.}
Another issue concerns scene adaptability. %~\cite{zhou2019learn}. 
%The indicator is given by joint model (see Table~\ref{tab:da}). 
Low joint model errors in both datasets have been a strong indicator that the pose regression adaptation is possible and the scene-invariant representations can accelerate the adaptation process. If, instead, the joint model error is higher, it can seriously compromise any chances of an efficient adaptation.
%A particular attention we pay to the cases of the low adaptability.
In the Cambridge Landmarks dataset, the joint models for the lately added Street scene indeed
%is already difficult in the supervised regression. Merging with any of four other scenes indeed 
show a high error. This suggests a low adaptability from and to Street scene. 
The adaptability theory developed an instance adaptation technique for low adaptability in classification tasks~\cite{chen19transferability}. Therefore it looks important to extend this theory to multi-regression tasks as well.

%===========================================
%\subsection{New discussion points}

%The performance of the regression layers has been longly criticized~\cite{zhou2019learn}. 

%exlude no 3D-based 

%In this paper we propose a different approach to generalization of the regression layers 
%When the transfer require one source scene to one target one source scene migth enout to tranefr

%Training on all but one (test) set
%Coupled with the retrieval 

%Still exist a margin of improvement . in 
%the scenario of aquisition a new datasets

%All previous methods required multiple scenes for training able to a new scene.

%improve in the 1-to-1 transfer scenario, transfer can be done

%Performance of 
%and generalize
%with low annotation cost

%Aquisition of new scenes from  massive non-annotated iamge 
%from weakly annotated scene with pose annotation. 

%Multi-source annotation

% Weak total non-annotated transfer 

%-----------------------------------
%\input{rebuttal/rebuttal.tex}
%==========================================
\section{Conclusion}
\label{sec:conclusion}

We address the problem of low generalization of the relative pose estimation models.
%Known for low generalization, the deep methods for absolute pose estimation. 
We attribute it to the dataset shift and propose adaptation across scenes as an alternative way towards a better generalization. We extend domain adaptation techniques invented for classification to the multi-regression task and developed a deep network to adapt a pose regression model from one scene to another. The Adversarial Pose Adaptation Network learns scene-invariant image representations and use target scene supplements to guide the transfer of source models to the target scene.
%We presented a adaptation for the problem of absolute pose estimation. 
%We consider the problem of representation transfer between two domains
%To address the output shift, we propose a semi-supervised approach when a tiny part of the target instances are annotated to guide the output transformation are considered the semi-supervised adaptation setting.
We also use the adaptability theory 
%to validate the existence of domain-invariant representation of images in two scenes and
to measure the transferability of feature representations from one scene to another.
%The domain we use the adversarial network to domain invariant representations.
We validate the superiority of the APANet on Cambridge Landmarks and 7Scene datasets over the baselines and compare them to the state of the art supervised methods.
%We also discussed alternative approaches to make the adaptation fast and efficient.

%----------------------------------

%\newpage
%{\small
\bibliographystyle{splncs04}
\bibliography{DomAd-all,bib-pose,bib-da}

\begin{thebibliography}{10}
\providecommand{\url}[1]{\texttt{#1}}
\providecommand{\urlprefix}{URL }
\providecommand{\doi}[1]{https://doi.org/#1}

\bibitem{balntas2018}
Balntas, V., Li, S., Prisacariu, V.: Relocnet: Continuous metric learning
  relocalisation using neural nets. In: European Conference Computer Vision
  (ECCV). pp. 782--799 (2018)

\bibitem{borchani2015survey}
Borchani, H., Varando, G., Bielza, C., Larra\~{n}aga, P.: A survey on
  multi-output regression. Wiley Int. Rev. Data Min. and Knowl. Disc.
  \textbf{5}(5),  216–233 (2015)

\bibitem{brachmann17dsac}
Brachmann, E., Krull, A., Nowozin, S., Shotton, J., Michel, F., Gumhold, S.,
  Rother, C.: {DSAC} - differentiable {RANSAC} for camera localization. In:
  Computer Vision Pattern Recognition (CVPR). pp. 2492--2500 (2017)

\bibitem{brachmann18learning}
Brachmann, E., Rother, C.: Learning less is more - 6d camera localization via
  3d surface regression. In: Computer Vision Pattern Recognition (CVPR). pp.
  4654--4662 (2018)

\bibitem{brahmbhatt18geometry}
Brahmbhatt, S., Gu, J., Kim, K., Hays, J., Kautz, J.: Geometry-aware learning
  of maps for camera localization. In: Computer Vision Pattern Recognition
  (CVPR). pp. 2616--2625 (2018)

\bibitem{cao2018-eccv}
Cao, Z., Ma, L., Long, M., Wang, J.: Partial adversarial domain adaptation. In:
  European Conference Computer Vision (ECCV). pp. 139--155 (2018)

\bibitem{cao2019learning}
Cao, Z., You, K., Long, M., Wang, J., Yang, Q.: Learning to transfer examples
  for partial domain adaptation. In: Computer Vision Pattern Recognition
  (CVPR). pp. 2985--2994 (2019)

\bibitem{chen16robust}
Chen, X., Monfort, M., Liu, A., Ziebart, B.D.: Robust covariate shift
  regression. In: Proc. {AISTATS}. vol.~51, pp. 1270--1279 (2016)

\bibitem{chen19transferability}
Chen, X., Wang, S., Long, M., Wang, J.: Transferability vs. discriminability:
  Batch spectral penalization for adversarial domain adaptation. In: Intern.
  Conference Machine Learning (ICML). vol.~97, pp. 1081--1090 (2019)

\bibitem{cortes2011domain}
Cortes, C., Mohri, M.: Domain adaptation in regression. In: Proc. 22nd Intern.
  Conf. on Algorithmic Learning Theory (2011)

\bibitem{ganin2016}
Ganin, Y., Ustinova, E., Ajakan, H., Germain, P., Larochelle, H., Laviolette,
  F., Marchand, M., Lempitsky, V.S.: Domain-adversarial training of neural
  networks. J. Mach. Learn. Res.  \textbf{17},  59:1--59:35 (2016)

\bibitem{gidaris2018unsupervised}
Gidaris, S., Singh, P., Komodakis, N.: Unsupervised representation learning by
  predicting image rotations. In: iclr (2018)

\bibitem{HoffmanX13Efficient}
Hoffman, J., Rodner, E., Donahue, J., Darrell, T., Saenko, K.: Efficient
  learning of domain-invariant image representations. CoRR
  \textbf{arXiv:1301.3224} (2013)

\bibitem{kendall16modelling}
Kendall, A., Cipolla, R.: Modelling uncertainty in deep learning for camera
  relocalization. In: {IEEE} International Conference on Robotics and
  Automation, {ICRA}. pp. 4762--4769 (2016)

\bibitem{kendall_geometric_2017}
Kendall, A., Cipolla, R.: Geometric {Loss} {Functions} for {Camera} {Pose}
  {Regression} with {Deep} {Learning}. In: Computer Vision Pattern Recognition
  (CVPR). pp. 6555--6564 (2017)

\bibitem{Kendall2015PoseNet:Relocalization}
Kendall, A., Grimes, M., Cipolla, R.: {PoseNet: A Convolutional Network for
  Real-Time 6-DOF Camera Relocalization}. In: Intern. Conference Computer
  Vision (ICCV). pp. 2938--2946 (2015)

\bibitem{kolesnikov2019revisiting}
Kolesnikov, A., Zhai, X., Beyer, L.: Revisiting self-supervised visual
  representation learning. In: cvpr. pp. 1920--1929 (2019)

\bibitem{laskar_camera_2017}
Laskar, Z., Melekhov, I., Kalia, S., Kannala, J.: Camera {Relocalization} by
  {Computing} {Pairwise} {Relative} {Poses} {Using} {Convolutional} {Neural}
  {Network}. In: {IEEE} International Conference on Computer Vision Workshops.
  pp. 929--938 (2017)

\bibitem{lathuiliere2018comrehensive}
Lathuili{\`{e}}re, S., Mesejo, P., Alameda{-}Pineda, X., Horaud, R.: A
  comprehensive analysis of deep regression. CoRR  \textbf{1803.08450} (2018)

\bibitem{leng2018local}
{Leng}, C., {Zhang}, H., {Li}, B., {Cai}, G., {Pei}, Z., {He}, L.: Local
  feature descriptor for image matching: A survey. IEEE Access  \textbf{7},
  6424--6434 (2019)

\bibitem{LongICML15Learning}
Long, M., Cao, Y., Wang, J., Jordan, M.I.: Learning transferable features with
  deep adaptation networks. In: Intern. Conference Machine Learning (ICML). pp.
  97--105 (2015)

\bibitem{LongCVPR14Transfer}
Long, M., Wang, J., Ding, G., Sun, J., Yu, P.S.: Transfer joint matching for
  unsupervised domain adaptation. In: Computer Vision Pattern Recognition
  (CVPR). pp. 1410--1417 (2014)

\bibitem{melekhov_relative_2017}
Melekhov, I., Ylioinas, J., Kannala, J., Rahtu, E.: Relative {Camera} {Pose}
  {Estimation} {Using} {Convolutional} {Neural} {Networks}. CoRR
  \textbf{1702.01381} (2017)

\bibitem{radwan2018vlocnet}
{Radwan}, N., {Valada}, A., {Burgard}, W.: Vlocnet++: Deep multitask learning
  for semantic visual localization and odometry. IEEE Robotics and Automation
  Letters  \textbf{3}(4),  4407--4414 (2018)

\bibitem{rublee11orb}
Rublee, E., Rabaud, V., Konolige, K., Bradski, G.R.: {ORB:} an efficient
  alternative to {SIFT} or {SURF}. In: Intern. Conference Computer Vision
  (ICCV). pp. 2564--2571 (2011)

\bibitem{saha18improved}
Saha, S., Varma, G., Jawahar, C.V.: Improved visual relocalization by
  discovering anchor points. In: British Machine Computer Vision (BMVC). p.~164
  (2018)

\bibitem{Saito_2018_ECCV}
Saito, K., Yamamoto, S., Ushiku, Y., Harada, T.: Open set domain adaptation by
  backpropagation. In: European Conference Computer Vision (ECCV). pp. 153--168
  (2018)

\bibitem{sattler_understanding_2019}
Sattler, T., Zhou, Q., Pollefeys, M., Leal-Taixe, L.: Understanding the
  {Limitations} of {CNN}-based {Absolute} {Camera} {Pose} {Regression}. In:
  Computer Vision Pattern Recognition (CVPR). pp. 3302--3312 (2019)

\bibitem{shavit_introduction2019}
Shavit, Y., Ferens, R.: Introduction to {Camera} {Pose} {Estimation} with
  {Deep} {Learning}. CoRR  \textbf{1907.05272} (2019)

\bibitem{shotton13scene}
Shotton, J., Glocker, B., Zach, C., Izadi, S., Criminisi, A., Fitzgibbon, A.W.:
  Scene coordinate regression forests for camera relocalization in {RGB-D}
  images. In: Computer Vision Pattern Recognition (CVPR). pp. 2930--2937 (2013)

\bibitem{torralba11unbiased}
Torralba, A., Efros, A.A.: Unbiased look at dataset bias. In: Computer Vision
  Pattern Recognition (CVPR). pp. 1521--1528 (2011)

\bibitem{valada18deep}
Valada, A., Radwan, N., Burgard, W.: Deep auxiliary learning for visual
  localization and odometry. In: {IEEE} International Conference on Robotics
  and Automation, {ICRA}. pp. 6939--6946 (2018)

\bibitem{walch17image}
Walch, F., Hazirbas, C., Leal{-}Taix{\'{e}}, L., Sattler, T., Hilsenbeck, S.,
  Cremers, D.: Image-based localization using lstms for structured feature
  correlation. In: Intern. Conference Computer Vision (ICCV). pp. 627--637
  (2017)

\bibitem{wang18deep}
Wang, M., Deng, W.: Deep visual domain adaptation: {A} survey. Neurocomputing
  \textbf{312},  135--153 (2018)

\bibitem{yang2019sanet}
Yang, L., Bai, Z., Tang, C., Li, H., Furukawa, Y., Tan, P.: Sanet: Scene
  agnostic network for camera localization. In: European Conference Computer
  Vision (ECCV). pp. 42--51 (2019)

\bibitem{you2019universal}
You, K., Long, M., Cao, Z., Wang, J., Jordan, M.I.: {Universal Domain
  Adaptation}. In: Computer Vision Pattern Recognition (CVPR). pp. 2720--2729
  (2019)

\bibitem{zhang17curriculum}
Zhang, Y., David, P., Gong, B.: Curriculum domain adaptation for semantic
  segmentation of urban scenes. In: Intern. Conference Computer Vision (ICCV).
  pp. 2039--2049 (2017)

\bibitem{zhou2019learn}
Zhou, Q., Sattler, T., Pollefeys, M., Leal-Taixe, L.: To learn or not to learn:
  Visual localization from essential matrices. CoRR  \textbf{abs/1908.01293}
  (2019)

\end{thebibliography}
%}

\end{document}